\documentclass{article}

\usepackage{microtype}
\usepackage{graphicx}
\usepackage{subfigure}
\usepackage{booktabs} 
\usepackage{placeins}

\usepackage{hyperref}

\usepackage[accepted]{icml2018}

\icmltitlerunning{Memory Augmented Self-Play}

\begin{document}

\twocolumn[
\icmltitle{Memory Augmented Self-Play}



\icmlsetsymbol{equal}{*}

\begin{icmlauthorlist}
\icmlauthor{Shagun Sodhani}{mila,equal}
\icmlauthor{Vardaan Pahuja}{mila,equal}
\end{icmlauthorlist}


\icmlaffiliation{mila}{MILA,  
    Universit\'e de Montr\'eal \\}

\icmlcorrespondingauthor{Shagun Sodhani}{sshagunsoddhani@gmail.com}
\icmlcorrespondingauthor{Vardaan Pahuja}{vardaanpahuja@gmail.com}

\icmlkeywords{Machine Learning, ICML}

\vskip 0.3in
]



\printAffiliationsAndNotice{\icmlEqualContribution} 

\begin{abstract}

Self-play ~\cite{2017arXiv170305407S} is an unsupervised training procedure which enables the reinforcement learning agents to explore the environment without requiring any external rewards. We augment the self-play setting by providing an external memory where the agent can store experience from the previous tasks. This enables the agent to come up with more diverse self-play tasks resulting in faster exploration of the environment. The agent pretrained in the memory augmented self-play setting easily outperforms the agent pretrained in no-memory self-play setting.

\end{abstract}

\section{Introduction}
\label{sec::introduction}

In a typical reinforcement learning task, the training agent has to find trajectories that lead to high rewards while learning to interact with the environment. Learning a good policy could require a large number of episodes as the agent has no understanding of the environment and only a small fraction of agent's interactions are useful for updating its policy, making these approaches sample inefficient. ~\cite{2017arXiv170305407S} presents an unsupervised training formulation, called as self-play, which enables the agent to explore and interact with its environment without using any external rewards. Two versions of the same agent, called as Alice and Bob, play a game where Alice proposes a task for Bob to perform. The task requires Bob to navigate through the environment and interact with different items thus enabling the agent to learn how to efficiently transition from one state to another. We use the word \textit{self-play} to refer to the technique proposed by ~\cite{2017arXiv170305407S}.

We extend the \textit{self-play} formulation and present a modified approach called as \textit{memory augmented self-play}. The key idea is to enable Alice to remember the tasks that she has assigned to Bob previously and use this experience while proposing a new task for Bob. We evaluate our approach on both discrete and continuous environments and present empirical evidence that the agents pre-trained with \textit{memory augmented self-play} outperform the agents pre-trained with \textit{self-play} for different tasks. We also show that \textit{memory augmented self-play} enables Alice to explore more parts of the environment than just \textit{self-play}.

\section{Motivation}
\label{sec::motivation}

\textit{Self-play} proposes a game between two variants of the same agent (referred to as Alice and Bob) where Alice assigns a task to Bob by performing some actions in the environment. The reward structure is such that the optimal behaviour for Alice is to come up with the simplest task that Bob can not finish and the optimal behaviour for Bob is to finish the given task as quickly as possible. Under the optimal setting, the series of tasks proposed by Alice would be such that each task is just slightly harder than the previous task. This would correspond to a curriculum for training Bob how to navigate through the environment and how to interact with the items in the environment. In literature, curriculum learning \cite{bengio2009curriculum} has been shown to boost the speed of convergence as well as enhance the generalization performance in the domain of supervised learning. In \cite{2017arXiv170705300F}, the agent learns to reach the goal from a set of start states increasingly far from the goal, which is another form of curriculum learning. The curriculum of start states depends on the performance of the agent at each step of the training process. Using this approach results in a significant performance gain in fine-grained robotic manipulation tasks.

While \textit{self-play} has been demonstrated to be an effective technique for unsupervised pre-training of the learning agents, there is a scope for improvement in some aspects of its formulation. Even though Alice is required to come up with a curriculum of tasks for Bob, she has no means of remembering what tasks she has already assigned to Bob. The reward signal tells her how hard the current task was for Bob but she has no explicit means of remembering this. This implies that Alice could assign similar tasks to Bob multiple times resulting in wasteful self-play episodes. Intuitively, the policy gradient algorithm has to learn how to remember the history of the previous tasks and how to use this history for generating a new task trajectory. 

We propose to overcome this limitation by providing an explicit memory module to Alice. In the \textit{self-play} setting, Alice is randomly initialized in the environment at the start of each episode and her actions are conditioned on her start state (for that episode) and her current state. The sequence of these actions generates the task for Bob. In the memory augmented setting, the actions that Alice take are conditioned on the memory as well as the start state and the current state in the episode. The memory is updated after every \textit{self-play} episode. This disentangling of memory and policy gradients allows policy gradients to generate more diverse trajectories which, in turn, help the agent to train faster.

\section{Model}

We use REINFORCE \cite{williams1992simple} algorithm with baseline for all the experiments. We first describe the agent architecture for \textit{self-play} and then we discuss the modifications made for \textit{memory augmented self-play}. We use a feature extraction network which is a single-layer feed-forward neural network. During an epsiode, at any step, the input to the network is a concatenated tuple of the current state in the current episode and the start state in the current episode (for Alice) or the current state in the current episode and target state in the current episode (for Bob). We refer to this tuple as the \textit{current episodic tuple} as the input. The network produces a feature vector of size \textit{feature-dim} using the \textit{current episodic tuple}. This feature vector is fed as an input to both the actor and the critic networks. The actor and critic networks are also single-layer feed-forward networks. In case of actor, the feed-forward network is followed by the softmax operator and in case of critic, we obtain a real-valued, scalar output.

The architecture of the feature extractor, actor network and the critic network are the same for all the environments. We used ReLU \cite{glorot2011deep} as the activation function.

\begin{table}[!ht]
\caption{Value of \textit{feature-dim} for different settings}
\label{shared_feat_table}
\centering
\resizebox{.45\textwidth}{!}{%
\begin{tabular}{|c|c|c|c|}
\hline
\textbf{Environment} & \textbf{Alice (self-play)} & \textbf{Alice (Memory)} & \textbf{Bob} \\ \hline
\textbf{Mazebase}    & 50                         & 100                     & 50           \\ \hline
\textbf{Acrobot}     & 10                         & 20                      & 10           \\ \hline
\end{tabular}
}
\end{table}

Table~\ref{shared_feat_table} summarizes the value of \textit{feature-dim} for the different settings. In all the cases, the size of the softmax layer is given by the number of possible actions in the environment except in the case of self-play episodes for Alice where she has an additional ``stop'' action. 

In the case of \textit{memory augmented self-play}, we use the \textit{current episodic tuple} as before to obtain the feature vectors. In the case of Alice, these features are concatenated with the features coming from the memory and the concatenated features are fed into the actor and the critic networks. Since Bob does not use the external memory, the architecture for Bob remains exactly the same as before. Now we discuss how the memory module is implemented and updated.  

We compared 3 different approaches to model the memory

\begin{enumerate}
    \item Last Episode Memory - In this setting, we obtain the feature vector corresponding to the \textit{last episodic tuple} and retain only that in the memory.
    \item Last $k$ Episode Memory - This setting generalizes the Last Episode Memory setting where we retain the feature vectors corresponding to the last $k$ episodes and average them to obtain the memory features.
    \item LSTM Episode Memory - In this setting, we pass the previous episodic feature vectors through a LSTM \cite{hochreiter1997long} and the hidden state of the LSTM gives the feature vector corresponding to the memory.
\end{enumerate}

Based on our experiments, we observed that LSTM Episode Memory performs better than the other two variants, so we used LSTM based memory for rest of the experiments.

\section{Experiments}
\label{sec::experiments}

We use PyTorch (0.3.1)\cite{paszke2017pytorch} as the framework for our implementation. The code for this implementation is available at the link \url{https://github.com/shagunsodhani/memory-augmented-self-play}. We evaluate our proposed approach on Mazebase ~\cite{2015arXiv151107401S} (grid world environment) and Acrobot. Mazebase is a discrete environment with discrete actions while Acrobot is a continuous environment with discrete actions. For running the Mazebase experiments, we used mazes of size ($8\times8$). The OpenAI Gym \cite{brockman2016openai} version of Acrobot environment was used for our experiments. For Mazebase, different models were trained till the current episodic reward saturated and for Acrobot, the experiments were run for 50K episodes. For all the reward curves and tables, we report rewards averaged over last $k$ episodes where $k$ is set to be 10000 for Mazebase and 2000 for Acrobot. Using other values of $k$ gives the same trend. The batch size for Mazebase was set to be 256 and that for Acrobot was 1 (using bigger batch size gives worse results). We used Adam \cite{kingma2014adam} optimizer (with learning rate of 0.001) and policy-gradient algorithm with baseline to train the agents. For self-play based experiments, we interleaved the training of self-play task and target task with one batch of self-play training followed by $n$ batches of training with the target task. $n$ is set to be $4$ for Mazebase and $100$ for Acrobot. For Mazebase, we set we set the maximum number of steps per episode, in the target task, to be 50, while the maximum number of steps, in the \textit{self-play} episode, is set to 80. In case of Acrobot task, we set the maximum number of steps per episode to 1000, while the maximum number of steps per \textit{self-play} episode is set to 2000.

\section{Results}
\subsection{Mazebase}
\begin{figure}[!ht]
\begin{center}
\centerline{\includegraphics[width=\columnwidth]{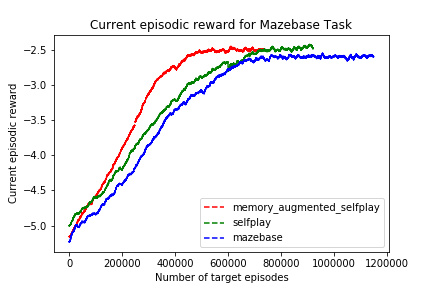}}
\caption{Figure showing the current episodic reward for the Mazebase Task using different strategies.}
\label{fig::mazebase_all}
\end{center}
\vskip -0.2in
\end{figure}

Figure ~\ref{fig::mazebase_all} shows how the trend in the value of current episodic reward (computed as a running average over last 10000 episodes) changes with the number of episodes for different tasks. We plot the different curves to the point of convergence and hence the three curves have different span along x-axis. The values of current episodic reward for some regularly sampled points are shown in Table ~\ref{tab::mazebase}.
The key observation is that \textit{memory augmented self-play} consistently performs better than both \textit{self-play} and \textit{no self-play} setting. Even though all the models converge towards the same reward value, using \textit{memory augmented self-play} increases the speed of convergence. All the curves were plotted using the average value after running the models for 5 different seeds. Figures ~\ref{fig::mazebase_mem}, ~\ref{fig::mazebase_selfplay}, and ~\ref{fig::mazebase_vanilla} (in the appendix) show the corresponding plots individually, along with the band of standard deviation. We also observe that \textit{self-play} (without memory) performs better than the no self-play setting. This was observed in the \textit{self-play} paper and our experiments validate that observation.

\subsection{Acrobot}

\begin{figure}[!ht]
\centering
\resizebox{.45\textwidth}{!}{\includegraphics[scale=0.7]{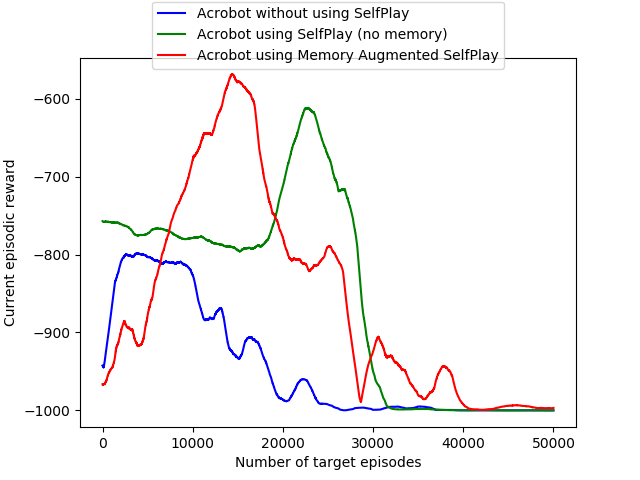}}
\caption{Figure showing the current episodic reward for the Acrobot Task using different strategies.}
\label{acrobot_comb}
\end{figure}

Figure ~\ref{acrobot_comb} shows the comparison of trend of current episodic reward (computed as a running average over last 2000 episodes) for the Acrobot task. All the curves were plotted using the average value after running the models for 3 seeds. Figures ~\ref{acrobot_selfplay_lstm}, ~\ref{acrobot_selfplay} and ~\ref{acrobot_noselfplay} (in the appendix) show the trend of current episodic reward (averaged over last 2000 episodes) with the number of episodes for the Acrobot task in \textit{memory augmented self-play}, \textit{self-play} and \textit{no self-play}, settings respectively. The values of average episodic reward for some regularly sampled points are shown in Table \ref{tab::acrobot}. For the Acrobot task, we observe that the memory augmented agent performs the best with the peak reward of -568.80. Even though both the \textit{self-play} and \textit{no self-play} variants perform poorly, the \textit{self-play} version is comparatively better with the peak reward of -611.72, as compared to -798.21 obtained with \textit{no self-play}.

\begin{figure}[!ht]
\centering
\resizebox{.45\textwidth}{!}{\includegraphics[scale=0.7]{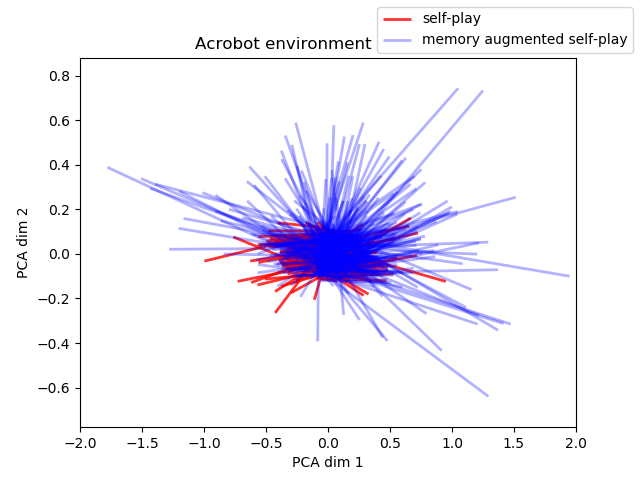}}
\caption{Plot of start and end states in 2D with and without memory augmentation}
\label{pca}
\end{figure}

One of our motivations behind adding memory was that having an explicit memory would allow Alice to remember what states she has already visited. This would enable her to explore more parts of the environment as compared to the \textit{self-play} setting. To validate this hypothesis, we perform a simple analysis. We compile the list of all the start and the end states that Alice encounters during training. Even though the start states are chosen randomly, the end states are determined by the actions she takes. We embed all the states into a 2-dimensional space using PCA and plot the line segments connecting the start and the end states for each episode. The resulting plot is shown in Figure \ref{pca}. We observe that using LSTM memory results in a wider distribution of end state points as compared to the case of \textit{self-play} with no memory. The mean euclidean distance between start and end points (in PCA space) increases from 0.0192 (self-play without memory) to 0.1079 (\textit{memory augmented self-play}), a $5\times$ improvement. This affirms our hypothesis that \textit{memory augmented self-play} enables Alice to explore more parts of the environment and enables her to come up with more diverse set of tasks for Bob to train on.

\begin{table}[!ht]
\centering
\caption{Comparison of trend of current episodic reward (averaged over last 10000 episodes) for different approaches on Mazebase Task}
\resizebox{.45\textwidth}{!}{%
\begin{tabular}{|c|c|c|c|}
\hline
\textbf{No. of episodes} & \textbf{No self-play} & \textbf{self-play} & \textbf{\begin{tabular}[c]{@{}l@{}}memory-augmented\\ self-play\end{tabular}} \\ \hline
\textbf{100K}             & -4.841               & -4.650            & -4.580                             \\ \hline
\textbf{200K}             & -4.419               & -4.253            & -3.899                             \\ \hline
\textbf{300K}             & -3.861               & -3.686            & -3.130                             \\ \hline
\textbf{400K}             & -3.357               & -3.265            & -2.758                             \\ \hline
\textbf{500K}             & -3.076                 & -2.942              & -2.541                               \\ \hline
\textbf{600K}             & -2.782                 & -2.802              &  -2.528                              \\ \hline
\textbf{700K}             & -2.669                 & -2.564              & -2.516                               \\ \hline
\end{tabular}
}
\label{tab::mazebase}
\end{table}


\begin{table}[!ht]
\centering
\caption{Comparison of trend of current episodic reward (averaged over last 2000 episodes) for different approaches on Acrobot task}
\resizebox{.45\textwidth}{!}{%
\begin{tabular}{|c|c|c|c|}
\hline
\textbf{No. of episodes} & \textbf{No self-play} & \textbf{self-play} & \textbf{\begin{tabular}[c]{@{}l@{}}memory-augmented\\ self-play\end{tabular}} \\ \hline
\textbf{10K}             & -826.31               & -778.74            & -678.14                             \\ \hline
\textbf{20K}             & -986.14               & -712.65            & -778.31                             \\ \hline
\textbf{30K}             & -999.24               & -949.59            & -924.37                             \\ \hline
\textbf{40K}             & -999.98               & -999.79            & -992.29                             \\ \hline
\textbf{50K}             & -1000                 & -1000              & -996.77                             \\ \hline
\end{tabular}
}
\label{tab::acrobot}
\end{table}


\FloatBarrier

\subsection{Conclusion}

We propose to augment the ~\textit{self-play} setting by providing a memory module to Alice so that she can remember the tasks she has assigned to Bob so far. We perform experiments on both discrete and continuous environments and provide empirical evidence that using \textit{memory augmented self-play} enables the agent to come up with more diverse self-play tasks. This leads to faster exploration of the environment and the agent pretrained in the \textit{memory augmented self-play} takes lesser time to train.

There are several interesting directions to extend this work. We have limited our experiments to use of a relatively simpler memory module. Further, we store just the start and the end state of each episode. A straight forward extension would be the have a hierarchical memory model where one level of memory is used to persist states across an episode and the second level of memory is used to maintain information across episodes. Another interesting extension would be to provide Alice with a  write-able differential memory. This would allow her to choose states that she wants to persist across episodes and may lead to a more efficient utilization of the memory. 

\FloatBarrier





\bibliography{main}

\begin{thebibliography}{10}
\providecommand{\natexlab}[1]{#1}
\providecommand{\url}[1]{\texttt{#1}}
\expandafter\ifx\csname urlstyle\endcsname\relax
  \providecommand{\doi}[1]{doi: #1}\else
  \providecommand{\doi}{doi: \begingroup \urlstyle{rm}\Url}\fi

\bibitem[Bengio et~al.(2009)Bengio, Louradour, Collobert, and
  Weston]{bengio2009curriculum}
Bengio, Yoshua, Louradour, J{\'e}r{\^o}me, Collobert, Ronan, and Weston, Jason.
\newblock Curriculum learning.
\newblock In \emph{Proceedings of the 26th annual international conference on
  machine learning}, pp.\  41--48. ACM, 2009.

\bibitem[Brockman et~al.(2016)Brockman, Cheung, Pettersson, Schneider,
  Schulman, Tang, and Zaremba]{brockman2016openai}
Brockman, Greg, Cheung, Vicki, Pettersson, Ludwig, Schneider, Jonas, Schulman,
  John, Tang, Jie, and Zaremba, Wojciech.
\newblock Openai gym.
\newblock \emph{arXiv preprint arXiv:1606.01540}, 2016.

\bibitem[{Florensa} et~al.(2017){Florensa}, {Held}, {Wulfmeier}, {Zhang}, and
  {Abbeel}]{2017arXiv170705300F}
{Florensa}, C., {Held}, D., {Wulfmeier}, M., {Zhang}, M., and {Abbeel}, P.
\newblock {Reverse Curriculum Generation for Reinforcement Learning}.
\newblock \emph{ArXiv e-prints}, July 2017.

\bibitem[Glorot et~al.(2011)Glorot, Bordes, and Bengio]{glorot2011deep}
Glorot, Xavier, Bordes, Antoine, and Bengio, Yoshua.
\newblock Deep sparse rectifier neural networks.
\newblock In \emph{Proceedings of the Fourteenth International Conference on
  Artificial Intelligence and Statistics}, pp.\  315--323, 2011.

\bibitem[Hochreiter \& Schmidhuber(1997)Hochreiter and
  Schmidhuber]{hochreiter1997long}
Hochreiter, Sepp and Schmidhuber, J{\"u}rgen.
\newblock Long short-term memory.
\newblock \emph{Neural computation}, 9\penalty0 (8):\penalty0 1735--1780, 1997.

\bibitem[Kingma \& Ba(2014)Kingma and Ba]{kingma2014adam}
Kingma, Diederik~P and Ba, Jimmy.
\newblock Adam: A method for stochastic optimization.
\newblock \emph{arXiv preprint arXiv:1412.6980}, 2014.

\bibitem[Paszke et~al.(2017)Paszke, Gross, Chintala, and
  Chanan]{paszke2017pytorch}
Paszke, Adam, Gross, Sam, Chintala, Soumith, and Chanan, Gregory.
\newblock Pytorch, 2017.

\bibitem[{Sukhbaatar} et~al.(2015){Sukhbaatar}, {Szlam}, {Synnaeve},
  {Chintala}, and {Fergus}]{2015arXiv151107401S}
{Sukhbaatar}, S., {Szlam}, A., {Synnaeve}, G., {Chintala}, S., and {Fergus}, R.
\newblock {MazeBase: A Sandbox for Learning from Games}.
\newblock \emph{ArXiv e-prints}, November 2015.

\bibitem[{Sukhbaatar} et~al.(2017){Sukhbaatar}, {Lin}, {Kostrikov}, {Synnaeve},
  {Szlam}, and {Fergus}]{2017arXiv170305407S}
{Sukhbaatar}, S., {Lin}, Z., {Kostrikov}, I., {Synnaeve}, G., {Szlam}, A., and
  {Fergus}, R.
\newblock {Intrinsic Motivation and Automatic Curricula via Asymmetric
  Self-Play}.
\newblock \emph{ArXiv e-prints}, March 2017.

\bibitem[Williams(1992)]{williams1992simple}
Williams, Ronald~J.
\newblock Simple statistical gradient-following algorithms for connectionist
  reinforcement learning.
\newblock In \emph{Reinforcement Learning}, pp.\  5--32. Springer, 1992.

\end{thebibliography}
\bibliographystyle{icml2018}

\newpage

\section{Appendix}

\begin{figure}[!ht]
\begin{center}
\centerline{\includegraphics[width=\columnwidth]{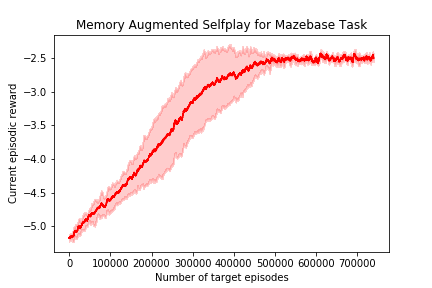}}
\caption{Figure showing the current episodic reward for the Mazebase Task using Memory Augmented SelfPlay.}
\label{fig::mazebase_mem}
\end{center}
\vskip -0.2in
\end{figure}

\begin{figure}[!ht]
\begin{center}
\centerline{\includegraphics[width=\columnwidth]{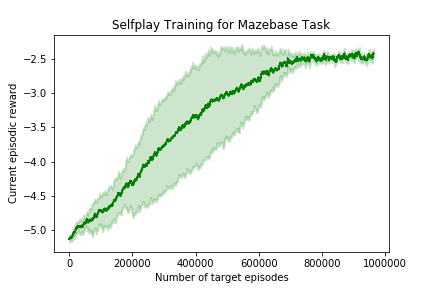}}
\caption{Figure showing the current episodic reward for the Mazebase Task using SelfPlay(no memory).}
\label{fig::mazebase_selfplay}
\end{center}
\vskip -0.2in
\end{figure}

\begin{figure}[!ht]
\begin{center}
\centerline{\includegraphics[width=\columnwidth]{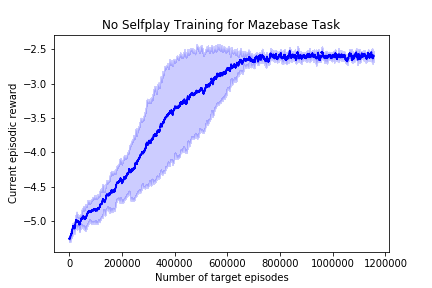}}
\caption{Figure showing the current episodic reward for the Mazebase Task without using SelfPlay.}
\label{fig::mazebase_vanilla}
\end{center}
\vskip -0.2in
\end{figure}

\begin{figure}[!ht]
\vskip 0.2in
\centering
\resizebox{.45\textwidth}{!}{\includegraphics[scale=0.7]{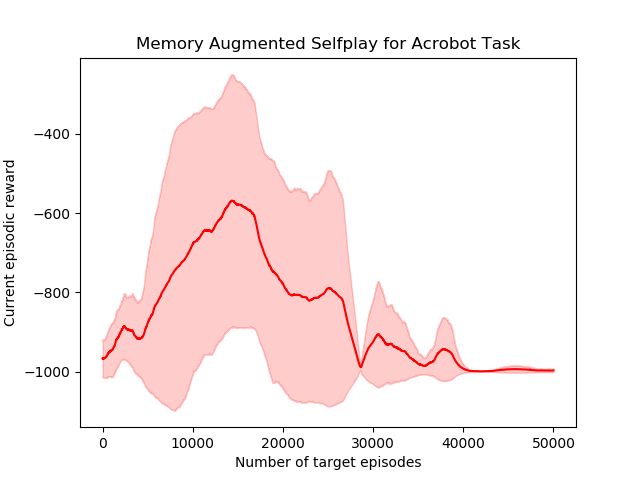}}
\caption{Figure showing the current episodic reward for the Acrobot Task using Memory Augmented SelfPlay.}
\label{acrobot_selfplay_lstm}
\end{figure}

\begin{figure}[!ht]
\centering
\resizebox{.45\textwidth}{!}{\includegraphics[scale=0.7]{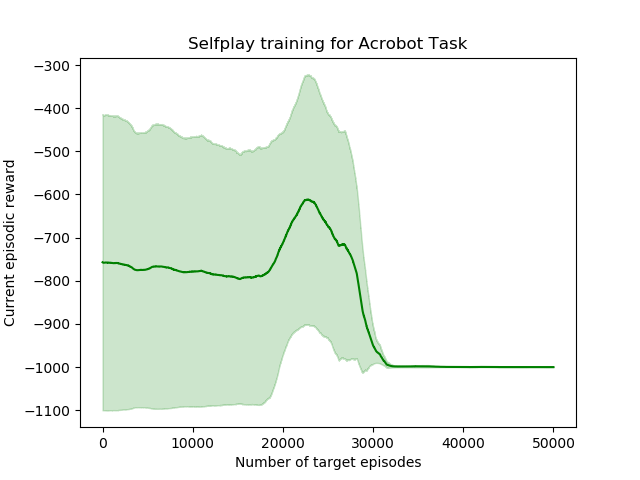}}
\caption{Figure showing the current episodic reward for the Acrobot Task using SelfPlay(no memory).}
\label{acrobot_selfplay}
\end{figure}

\begin{figure}[!ht]
\centering
\resizebox{.45\textwidth}{!}{\includegraphics[scale=0.7]{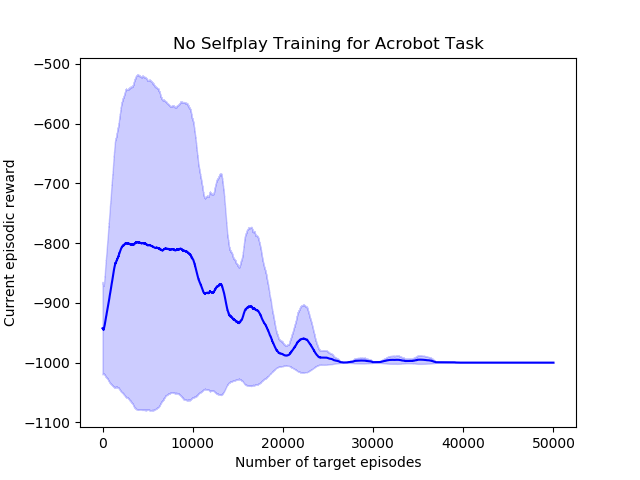}}
\caption{Figure showing the current episodic reward for the Acrobot Task without using SelfPlay.}
\label{acrobot_noselfplay}
\end{figure}





\end{document}